\documentclass[journal]{IEEEtran}
\usepackage{graphicx}
\usepackage{booktabs}
\usepackage{subcaption}
\usepackage{amssymb}
\usepackage{multirow}
\usepackage{dsfont }
\usepackage{xspace}
\usepackage{amsmath}

\ifCLASSINFOpdf
\else
\fi
\begin{document}
\title{Cross-Image Region Mining with Region Prototypical Network for Weakly Supervised Segmentation}

\author{Weide~Liu,
        Xiangfei~Kong,
        Tzu-Yi~Hung,
        Guosheng~Lin

\thanks{Weide Liu is with School of Computer Science and Engineering, Nanyang Technological University (NTU), Singapore 639798 (e-mail: weide001@e.ntu.edu.sg) and with the Institute for Infocomm Research (I$^2$R)-Agency for Science, Technology and Research (A*star), Singapore, 138632.}

\thanks{X. Kong is with Ant Group, Singapore (e-mail: xiangfei.kong@antgroup.com).}

\thanks{T.~Hung is with Delta Research Center, Singapore (e-mail:  tzuyi.hung@deltaww.com).}

\thanks{G. Lin is with School of Computer Science and Engineering, Nanyang Technological University (NTU), Singapore 639798 (e-mail: gslin@ntu.edu.sg).}

\thanks{The code is available at https://github.com/liuweide01/RPNet-Weakly-Supervised-Segmentation}

\thanks{Corresponding author: Guosheng Lin.}
}

\markboth{IEEE Transactions on Multimedia}%
{Shell \MakeLowercase{\textit{et al.}}: Bare Demo of IEEEtran.cls for IEEE Journals}

\maketitle

\begin{abstract}
Weakly supervised image segmentation trained with image-level labels usually suffers from inaccurate coverage of object areas during the generation of the pseudo groundtruth. This is because the object activation maps are trained with the classification objective and lack the ability to generalize.
To improve the generality of the object activation maps, we propose a region prototypical network (\textbf{RPNet}) to explore the cross-image object diversity of the training set. Similar object parts across images are identified via region feature comparison. Object confidence is propagated between regions to discover new object areas while background regions are suppressed. Experiments show that the proposed method generates more complete and accurate pseudo object masks while achieving state-of-the-art performance on PASCAL VOC 2012 and MS COCO. In addition, we investigate the robustness of the proposed method on reduced training sets.
\end{abstract}


%
\IEEEpeerreviewmaketitle

\section{Introduction} \label{sec: introduction}
\label{sec:Introduction}
\IEEEPARstart{S}{emantic} segmentation is a task to assign each pixel in a scene image with a semantic category. One of the biggest challenges of this task is the numerous, time-prohibitive efforts entailed in the manual labeling of a large set of training images. 

To alleviate, or even free researchers from high costs of laborious annotations, more attention has been paid to weakly supervised semantic segmentation, in which annotations can be performed in a much-eased manner: rather than associating all pixels from an image with a label, weaker supervisions on the training images such as bounding boxes~\cite{box,box2}, scribbles~\cite{scribble}, image labels~\cite{irn,psa,ssdd,cvpr17web,cvpr18web,triplet,sec,cbts,seam,Pathak215,dcsm,meff,lee_iccv2019,tmm-zhang2019decoupled}, and points~\cite{point} can be used as substitute annotations to train.

Compared with other annotations, the image-level label is cost-friendly and readily accessible to the public. There is a broad choice of large image corpora with image-level labels such as ImageNet~\cite{imgnet}, PASCAL VOC~\cite{voc}, and images retrieved by search engines with their keywords as labels \cite{webvideo-seg}.
However, image-level labels merely provide categorical cues without any shape/texture information of any objects. It is crucial to utilize such cues to automatically produce coherent regions of objects with their pixel-level categorical information.
To this end, Class Activation Maps (CAM)~\cite{cam} has been wildly used to explore the localization and shape cues from image-level labels~\cite{irn,psa,ssdd,cvpr17web,cvpr18web,triplet,sec,cbts,seam}.
However, there is no guarantee of the completeness of the object activation regions of CAM since they are trained only with the classification objective. Incomplete or wrong activation maps obtained by CAM may lead to sub-optimal performance.

To improve the quality of the object activation maps, many approaches are proposed to use the results from CAM as the seed activation regions and gradually refine their quality. Ahn~\textit{et al.}~\cite{irn} use low-level image features to enhance the coherence of the activation maps. Wang~\textit{et al.}~\cite{seam} expand the seed object regions by fusing the object activation maps computed on different spatial scales of the input image. Object regions that are not activated by CAM at one scale could be re-activated at a coarser level. However, this method does not explore the rich cross-image context within the same class in a training set. 
CIAN~\cite{cian} addresses the cross-image context by computing an affinity map for each pixel of an image to another. However, the pixel of the query may not be the foreground and could end up retrieving the wrongly activated backgrounds.

Hu~\textit{et al.}~\cite{seenet} find the cross-image context leveraging two self-erasing networks with hiding and erasing strategy to explore more object regions. New regions will be activated by the classification objective if the old ones are masked. However, this method cannot automatically stop the erasing even after the entire foreground is masked, resulting in the undesirable background being activated.

\begin{figure*}[t]
 \centering
    \includegraphics[width=1\linewidth]{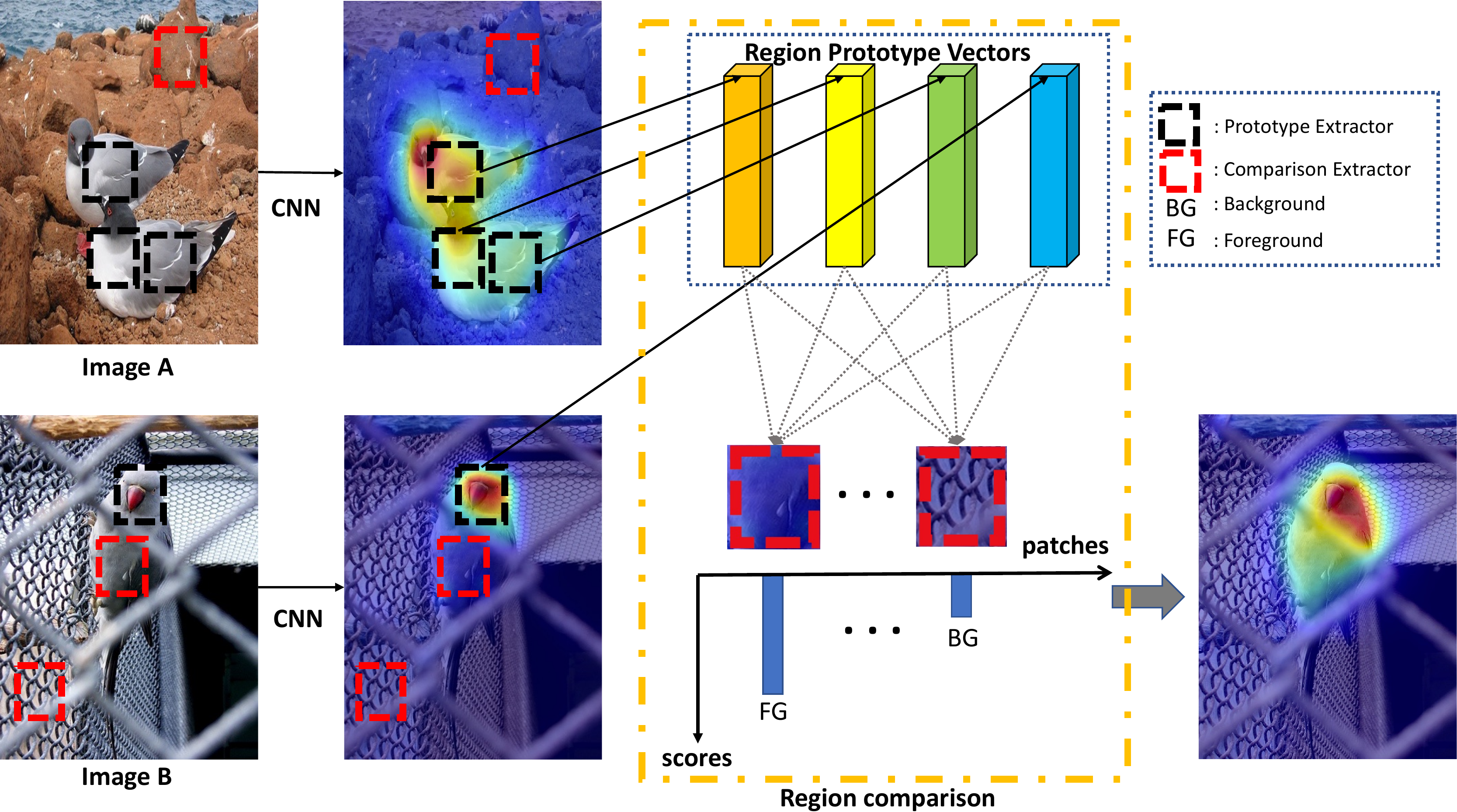}
    \caption{Motivation and Region comparison. Given a pair of sample images that contain common classes as the network input, we first encoder the images into object activation maps, and generate the region prototype vectors by collecting the confident object regions. By comparing the similarity to the region prototype vectors, we re-weight the inactive object regions. 
    }
    \label{Figure:motivation}
\end{figure*}

Inspired by the previous works, we propose that the key to improve the quality of the object activation maps is 1) a method to locate more complete object regions and 2) a mechanism to suppress wrongly activated background regions. We hold an assumption that the object regions possess significantly less diversity compared with the background. We use it to address the two key points:
To point 1), we find that multiple training samples of the same class provide a diversity of object regions, see Figure~\ref{Figure:motivation}. One activated object region in an image is helpful to explore more similar regions in another image of the same class. In particular, we generate the region prototypes to capture the diversity by collecting regions with similar features across the same object category images; then, we locate the inactivated object regions in these images by comparing their feature maps to the region prototypes. Suppose one prototype is extracted from image A; this prototype will be compared to all regions in image A and the pairing image B, including all regions of the other prototypes. Similar regions are highlighted and assigned with large confidence values as new object regions. 

To point 2), if an activated region belongs to the background, it is difficult to propagate its information, among others. Since there is more diversity among backgrounds, fewer compared regions during prototype voting resemble the activated background region, and its information propagation is minimized. Our region comparison serves as a voting mechanism to assign them with lower confidence scores.

With the region prototype mechanism on cross-images, our proposed network generates better object activation maps to train a segmentation model. We conduct extensive experiments on PASCAL VOC 2012~\cite{voc} and MS COCO \cite{coco14} to validate the effectiveness of our network. However, these datasets both have abundant training samples and do not cover a practical scenario in which fewer training samples are available. A good method should produce a high-quality pseudo groundtruth even when the training samples are scarce. Thus, we investigate the robustness of the proposed method on datasets of reduced training samples (\textit{e.g.} up to $1/16$ of the number of training samples in PASCAL). Our main contributions are summarized as follows:

\begin{itemize}
	\item  The proposed extraction of region prototypes and their equal treatments ensure the reactivation of rare object regions that are not activated by CAM while preventing them from being dominated by prevalent features. 
	\item The proposed voting mechanism on prototypes propagates important object information while suppressing backgrounds simultaneously, bringing in more robustness to either abundant or limited training images.
    \item We are the first to evaluate weakly supervised segmentation methods under limited training samples and in different backbones. The proposed method exhibits promising results under both conditions.
    \item We improve the performances on PASCAL VOC 2012 and MS COCO dataset and achieve new state-of-the-art.
\end{itemize}

 \begin{figure*}[t]
    \centering
    \includegraphics[width=1\linewidth]{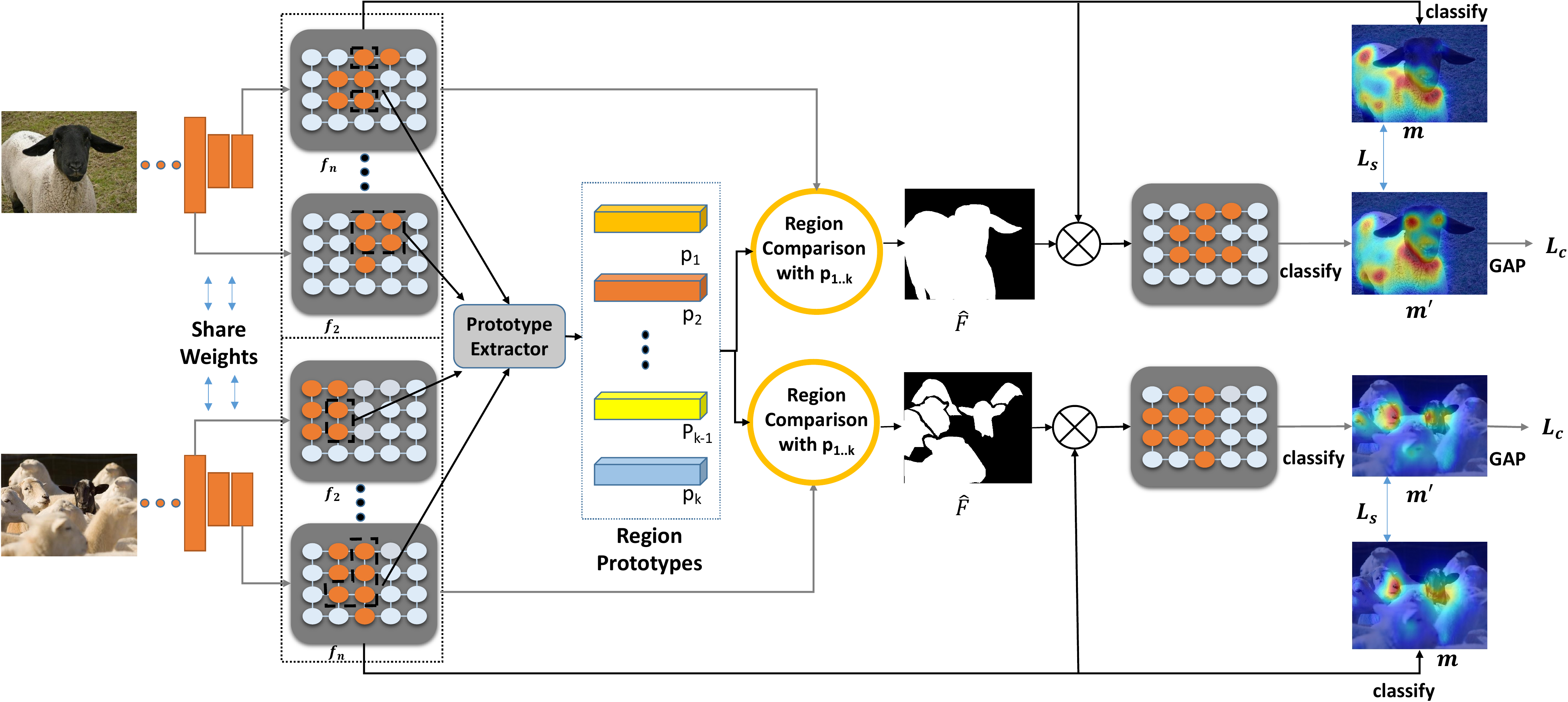}
    \caption{The architecture of our methods. Given a pair of sample images that contain common classes as the network input, we first generate object activation maps with a parameter-shared encoder, then we reinforce the representations on the object-related regions, which are explored by region prototype vectors. We select different object-related regions of multiple-level features to generate the region prototype vectors and use them to generate the foreground probability maps. 
    Besides the standard multi-label classification loss $\mathcal{L}_c$, we also add an auxiliary loss $\mathcal{L}_{s}$ that uses the refined prediction to supervise the training of the original prediction. }
    \label{Figure:overall}
\end{figure*}

 \begin{figure}[t]
    \centering
    \includegraphics[width=1\linewidth]{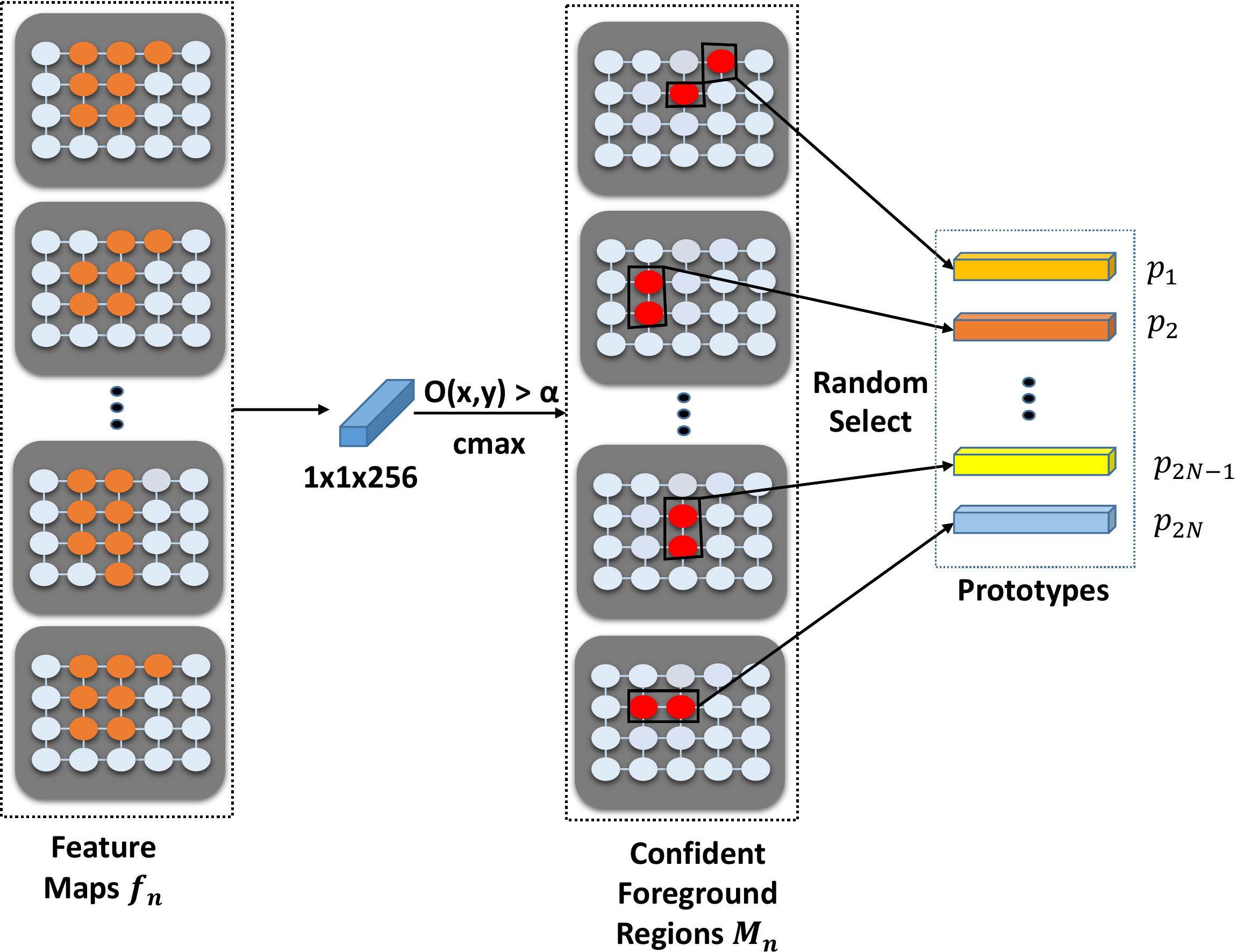}
    \caption{The architecture of our prototype extractor}
    \label{Figure:protype-extractor}
\end{figure}
\section{Related Work}
\subsection{Semantic segmentation}
Semantic segmentation is a fundamental computer vision task that assigns each pixel in the image with a category. Currently, state-of-the-art methods handle image semantic segmentation as a dense prediction task and adopt fully convolutional networks to make predictions~\cite{chen2018deeplab,long2015fully}. To make pixel-level dense predictions, encoder-decoder structures~\cite{tip-encoder-decoder1,tip-encoder-decoder2,tmm-kang2018depth,tmm-shi2018hierarchical,tmm-zhang2019decoupled,lwd2020crnet,lwd2020guided,lwd2021few,liu2021few_transport,lwd2022long,hou2022distilling} are widely used to reconstruct high-resolution prediction maps. Typically, an encoder gradually downsamples the feature maps, acquiring a large field-of-view and capturing the semantic object information. Then, the decoder gradually recovers the fine-grained information.

The field-of-view information is important for semantic segmentation tasks. Dilation connections~\cite{dilation} are often used to increase the field-of-view and then fuse high-level and low-level features for better predictions. We also follow the encoder-decoder design in our network and opt to transfer the guidance information in the low-resolution maps and use decoders to recover details.

\subsection{Weakly supervised semantic segmentation.}
To alleviate the data deficiency problem in image segmentation task, the weak supervision have been explored, such as supervision with bounding boxes~\cite{box,box2}, scribbles~\cite{scribble}, and image labels~\cite{irn,psa,ssdd,cvpr17web,cvpr18web,triplet,sec,cbts,Pathak215,dcsm,meff,seam,benet}.
The weak supervision methods have the nature of less expensive to be annotated and easily acquired. 

Previous state-of-the-art methods~\cite{irn,psa,erasing} mostly adopt CAMs~\cite{cam} to localize the objects. However, the CAMs can only locate the most discriminative object regions, which is insufficient to train a segmentation model.  Researchers have designed various ways to propagate the seed regions to the entire object to explore more object-relative regions. 
Hou~\textit{et al.}~\cite{seenet} expand the seed regions by erasing the currently discovered regions with prohibit attention and re-train the erased image.
SPM~\cite{spm} expands the seed regions with a discrepancy and intersection loss. Ahn~\textit{et al.}~\cite{irn} propagate the seed regions to reach the boundary and utilize the inter-pixel relationship to refine the final pseudo ground mask. 
CIAN~\cite{cian} retrieves the cross-image relation by calculating the dot product between every pixel from different images, which do not distinguish the foreground and background. In contrast to CIAN, we generate the region prototypes only from the confident object regions. 
CONTA~\cite{conta} proposes a structural causal model to remove the confounding bias in image-level classification. Our RPNet also can suppress the wrong foreground by a voting mechanism with multiple prototypes.

\begin{figure*}[t]
    \centering
    \includegraphics[width=1\linewidth]{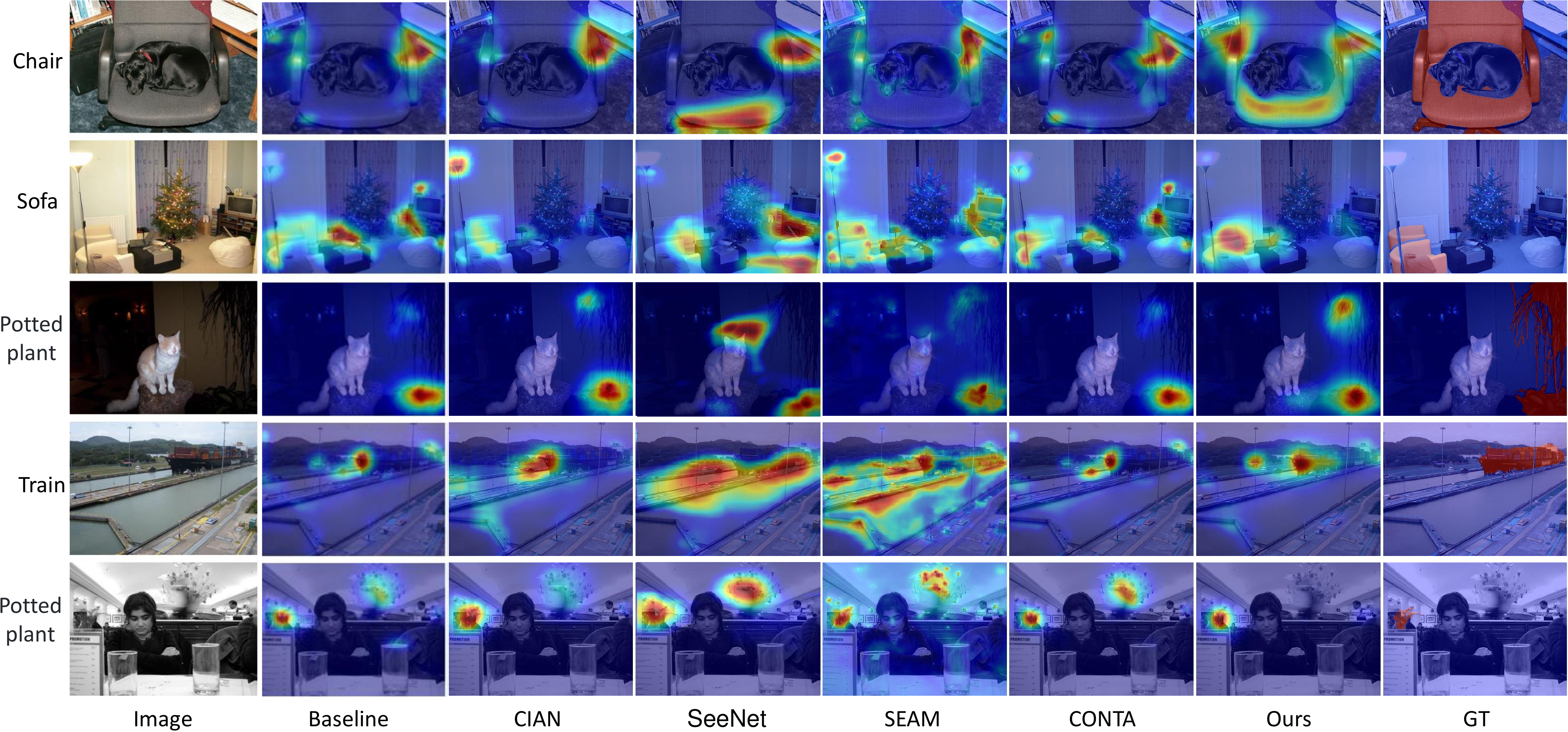}
    \caption{Visualization examples that compare the object activation maps generated by different methods: CIAN~\cite{cian}, SeeNet~\cite{seenet}, SEAM~\cite{seam}, CONTA~\cite{conta}, and Ours: RPNet, GT denotes the ground truth. As shown in the figures, our RPNet can accurately predict the object localization and cover most object regions. }
    \label{Figure:different-methods}
\vspace{+0.2cm}
\end{figure*}

Compared with previous methods, our method expands the seed region by leveraging the cross-image relations from a novel region prototype perspective to mine the inactive object regions and suppress the dislocated foreground.
SEAM\cite{seam} learns a network to enforce the classifier to recognize the common semantics from co-attentive objects. The cues of the foreground sometimes have to be refined by extra saliency information. In contrast, the proposed RPNet selects the confident object regions from different images to retrieve similar object parts while the background features can be suppressed simultaneously.
In particular, we further compare the pixel correlation module (PCM) proposed by SEAM~\cite{seam} to our RPNet, including:
1. The PCM in SEAM computes attention maps with simple pixel-to-pixel differences, which misses the neighboring semantic information. The proposed PR-Net addresses this problem by leveraging region-based prototypes to compute our attention maps. The difference in our attention maps comes from low-level image features vs. high-level semantic features.
2. The SEAM computes attention maps at the pixel level of its feature maps without considering whether this pixel belongs to the foreground or the background. As an example shown in Figure~\ref{Figure:different-methods}, if one background pixel on the feature maps is wrongly activated, other background pixels similar to this one will be re-activated as foreground through its correlation module, even though they were labeled as background before. Thus, in SEAM, background regions are more prone to be falsely activated. On the contrary, our RPNet carefully selects regions with high foreground confidence as prototypes before propagating their information to re-activate other unnoticed regions. The RPNet tries to guarantee that only correct (foreground) information is propagated.
3. SEAM's correlation module is confined within single images, while our RPNet uses cross-image to explore the diversity of the foreground features. More robustness and completeness of the object activation maps are expected in the outputs of the RPNet.

MCI\cite{mining-cross} learns a network to enforce the classifier to recognize the common semantics from co-attentive objects. The cues of background sometimes need to be refined by extra saliency information. Our RPNet selects the confident object regions from different images to simultaneously retrieve similar object parts while suppressing the background features.

IIC\cite{zhang2020interimage} and our proposed method both seek to re-activate object regions of the same object category in different images. However, this work maintains a memory bank, and the centers of prototypes are used, which could cause the prototypes to be dominated by some most popular object regions, e.g., memory centers could be severely biased to dog heads rather than dogs' whole bodies. This is perhaps less serious for detection tasks (which is the focus of \cite{zhang2020interimage}) but is very harmful to image segmentation.
\section{Proposed Method}
\subsection{Motivation}
\label{section:motivation}
The object activation maps of CAM are sub-optimal for fully supervised segmentation training because they are trained via a classification objective. We argue that such maps suffer from two types of flaws: 1)incomplete object regions (foreground) and 2) falsely activated cluttered regions (background). Generally, when the training set has abundant samples in each category, the trained model benefits from it and tends to be robust to backgrounds. However, at the same time, the activated object region tends to be more incomplete since some less distinctive parts tend to be suppressed by the classification objective, \textit{e.g.} the torso part of a mammal which is less distinctive compared with its head. This means flaw 1) is more likely to happen. When the training samples become scarce, flaw 2) happens more as some backgrounds could be discriminative with fewer data available to counter possible overfitting. See Figure \ref{Fig:decrease-data} and our observation in the experiment part in Sec. \ref{sec:decress_training} on truncated versions of \textit{PASCAL VOC 2012}. It is crucial for a method to be robust to both abundant or scarce training samples and address the two mentioned flaws. To our knowledge, we are the first to investigate the number of training samples and handle both flaws elegantly to produce high-quality object activation maps.~\cite{seam},\cite{cian} and~\cite{seenet} consider the flaw 1) only, while~\cite{conta} considers flaw 2) only.

\subsection{Method Pipeline}
As shown in Figure~\ref{Figure:overall}, given a training set of images with image-level labels, we train a model that automatically produces initial object regions. The model extracts object features similar to CAM~\cite{cam} but is from multiple spatial resolutions. Based on these features, we explore the diversity of the training images in each object class: at each iteration, we compute a group of cross-image region prototype vectors on a pair of sample images that contain a common object class. These prototypes are compared with the feature maps of the inputs and find out more inactive foreground regions. Two loss functions, namely classification loss, and self-supervised loss, are used to ensure the precision of the foreground activation and suppress the background regions during training. The trained model is used to predict maps of initial object regions on the input images with their image-level labels.

The proposed method can be integrated into the popular pipeline for weakly supervised segmentation training~\cite{irn,psa,ssdd,conta,seam}: After the initial object regions are computed, boundary refinement~\cite{irn} is applied on them to generate the pseudo groundtruth, on which the fully-supervised segmentation model can be trained.

\subsection{Revisiting Class Activation Maps}
The proposed method starts with the results of CAM~\cite{cam} and explores useful cross-image context based on them. CAM computes the object activation maps via a classification network backbone connected with a global average pooling (\textbf{GAP}) layer as follows:
\begin{equation}
    A(x,y) =g( \frac{\Theta _{}^{T} f_n(x,y) }  { \max (\Theta _{}^{T} f_n(x,y) )} ),
    \label{equation_cam}
\end{equation}
\begin{equation}
    CAM(x,y) = A(x,y) \cdot c,
\end{equation}
where $\textit{$f_n$} \in \mathbb{R}^{X \times Y \times D}$ denotes the feature maps extracted from the last convolution block. $(x,y)$ denote 2D coordinates, $\Theta$ is the classification weights and $g$ is the is the activation function, which is $ReLU$ in this paper. The $X, Y$ are the feature sizes while the $D$ is the channel dimension.
The spatially normalized object activation map is denoted as $A \in \mathbb{R}^{X \times Y \times C}$ and the per-class object activation map $CAM(x,y)$ is computed by masking $A$ with the image-level label vector $c$ containing total $C$ classes.

 \begin{figure}[t]
    \centering
    \includegraphics[width=1\linewidth]{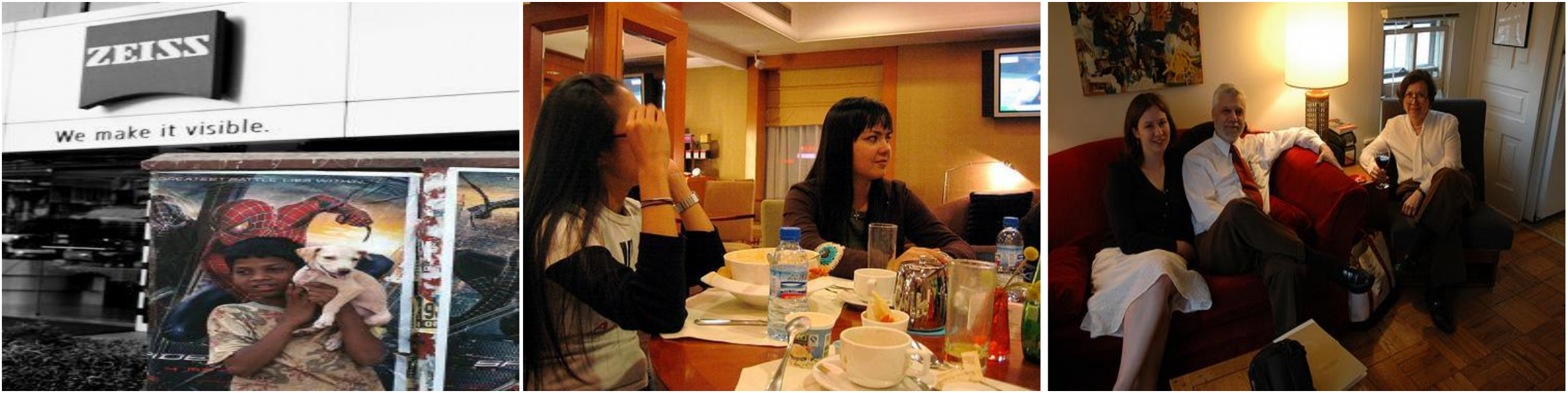}
    \caption{ We randomly sample some pictures from the VOC 2012, the foreground contains `person, table, sofa.' However, the background includes `fresco, cartoon, house, window, TV, food, cup, trademark...' In one training iteration, e.g., if the object class ``person'' is selected as the foreground, while large parts of cluttered background regions with more diversified features (compared with features of ``person'') are compared. Our assumption is based on the diversity comparison between ``single object category'' vs. all other objects and backgrounds instead of ``all object'' vs. backgrounds.}
    \label{Figure:background}
\end{figure}
\subsection{Region Prototypical Network}
\label{sec: rpnet}
We propose the region prototype that captures distinctive features among sample images of the same object category: prototypes extracted from different samples are compared with their feature maps. We define this step as \textbf{region comparison} as shown in Figure \ref{Figure:motivation}. We hold a simple assumption: the diversity of the object regions is \textbf{far less} than the diversity of background. 
e.g., we random sample some pictures from the VOC 2012, as shown in Figure~\ref{Figure:background}, the foreground contains `person, table, sofa.' However, the background includes `fresco, cartoon, grass, house, window, TV, food, cup, trademark...' In one training iteration, only the object class “person” is selected as the foreground, while large parts of cluttered background regions with more diversified features (compared with features of “person”) are compared. Our assumption is based on the diversity comparison between “single object category” vs. all other objects and backgrounds instead of “all objects” vs. backgrounds.

Thus, if an extracted prototype is truly an object region, with the help of the diversity of the training images, this prototype's information will propagate well during training and re-activate similar regions from other samples. On the contrary, if the prototype is background wrongly activated by the classification objective, the information propagation is much slower than object regions: the diversity of background is significantly more, and these prototypes can hardly find their similar neighbors. 
As the example shown in Figure~\ref{Figure:motivation}, only the bird's head is activated in Image B, while the head and body regions are activated in Image A. The body features in image A can be used to guide image B to find out its inactivated body regions and highlight them. Based on this finding, we build a region prototype network to extract cross-image object features as prototypes and take pairs of images within the same object category as inputs. This aims to have full utilization of the object diversity among the training set. We are not the first to use image pairs as inputs: \cite{cian} explores cross-image context by computing an affinity map for each pixel for image pairs.  

\textbf{Generate regions prototype vectors.} To fully explore the object information, we compute the normalized activation feature maps $O  \in \mathbb{R}^{X \times Y \times D}$ on multiple feature maps of the image pair, \textit{e.g.} $\{f_n(x,y)|n=1...N\}$, where $N$ is the index of resnet blocks in a classification network backbone. We unify of the number of channels by concatenation of each $f_n$ with a 256 channel $1\times1$ convolution layer:
\begin{equation}
    O_n(x,y) = conv\ast g(\frac{f_n(x,y)}{\max f_n(x,y)}).
    \label{equation:object_activation}
\end{equation}

Unlike the prototypical network~\cite{prototipical} which spatially averages the entire feature maps as the prototypes and introduces noise, our RPNet carefully selects activation regions as the region prototype vectors with high confidence only. Specifically, given a set of object activation maps $\{O_n(x,y)\}$, we compute their confident object activation maps $O_n^{'} \in \mathbb{R}^{X \times Y \times 1}$ by performing a maximum operation over the spatial dimension (\textbf{$cmax$}):
\begin{equation}
    O_n^{'}(x,y) = cmax(O_n(x,y)).
\end{equation} 

In equation \ref{equation:object_activation}, We do not use the image-level label $c$ as CAM does in equation \ref{equation_cam} to locate object regions, due to the inherent flaw of the classification objective previously discussed. Instead, all channels are given equal possibilities to be highlighted as the foreground to produce more complete object regions. We generate the region prototype vectors $\{p_n \in \mathbb{R}^{1 \times 1 \times D}\}$ as:

 \begin{figure*}[t]
  \centering
    \includegraphics[width=1\linewidth]{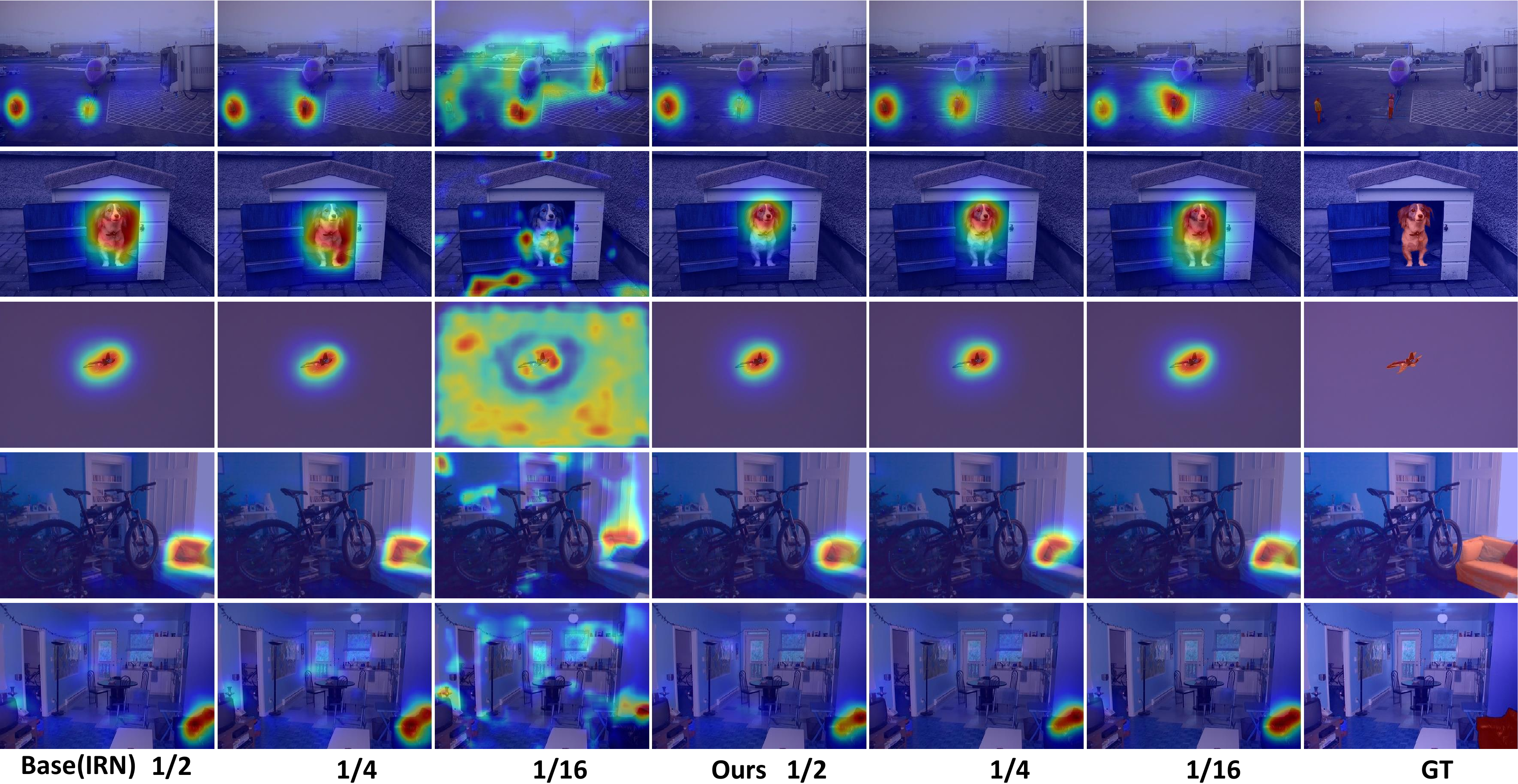}
    \caption{The object activation maps trained with different numbers of the PASCAL VOC 2012~\cite{voc} training data. 1/2 denotes using only half the training data, 1/4 denotes using only quarter training data, and so on. Our baseline is IRN~\cite{irn}. }
    \label{Fig:decrease-data}
\end{figure*}
 \begin{figure*}[t]
    \centering
    \includegraphics[width=1\linewidth]{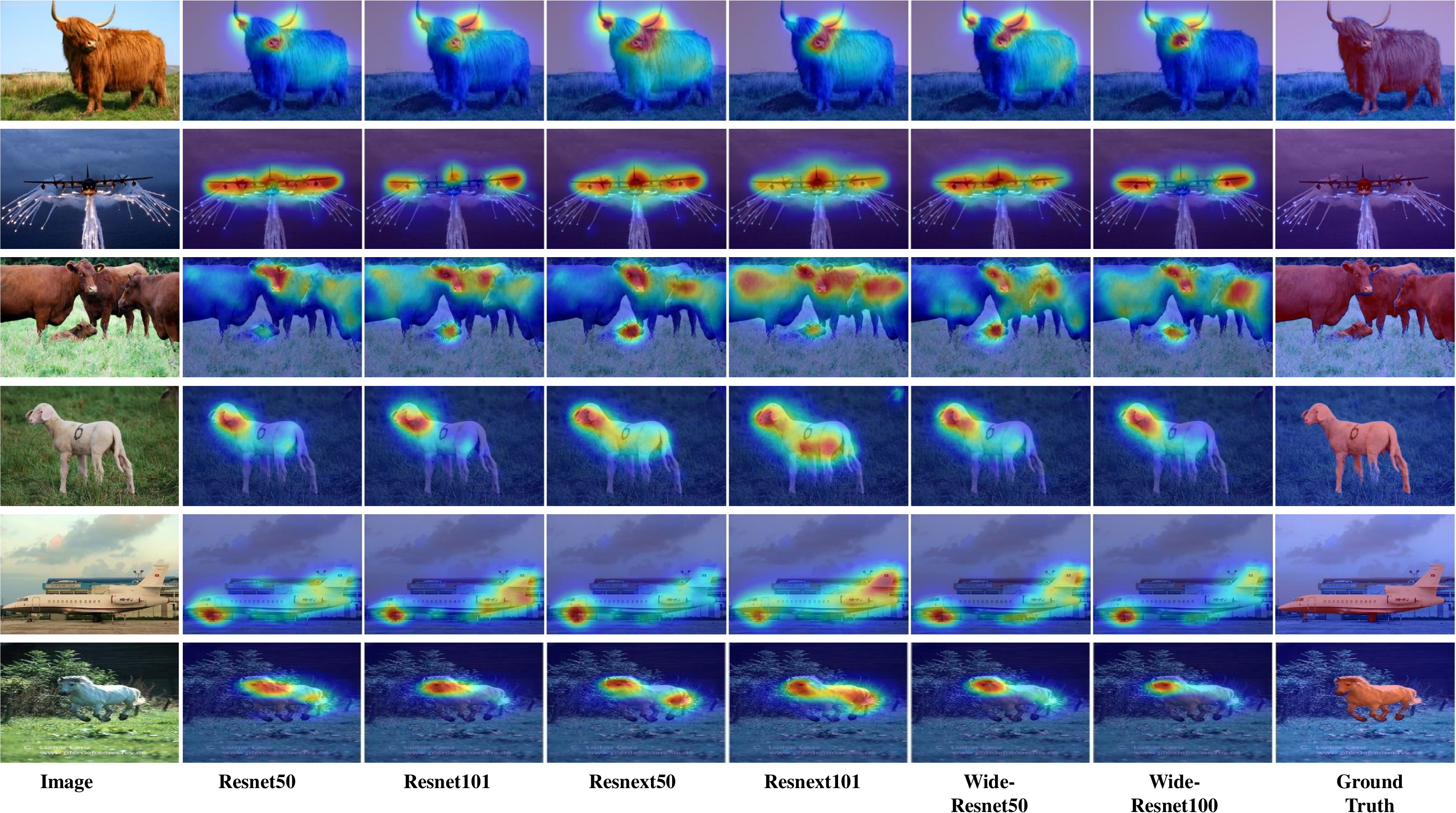}
    \caption{Visualization examples that compare the object activation maps generated by the different backbones. }
    \label{Figure:different-backbone}
\end{figure*}
\begin{table*}[t]
\centering
\small

\caption{The performance of the synthesized segmentation labels evaluated on the PASCAL VOC 2012 and MS COCO training set. Base: base object activation maps, RPNet: our enhanced object activation maps.  BR: Boundary Refine~\cite{irn}, CONTA~\cite{conta}. }

\begin{tabular}{c|c|c|c|c|c|c}
\toprule
Dataset & Base & Base + CONTA & Base + RPNet & Base + BR & Base + CONTA + BR & Base + RPNet + BR \\ \hline
VOC & 48.3  &48.8  & \textbf{50.8}  & 66.0     &67.9    & \textbf{69.0} \\ \hline
COCO & 27.4 &28.7  & \textbf{36.1}  & 33.9     &35.2    & \textbf{44.9} \\ 
\bottomrule
\end{tabular}

\label{psedo_mask}
\vspace{+0.2cm}
\end{table*}
\begin{table}[t]

\centering
\small

\caption{Comparison of different training data volume on PASCAL VOC 2012. 1 denotes use all the VOC 2012 training data; 1/2 denotes use half of the dataset, and so on.  When the training data decrease to 1/16, our RPNet can achieve comparable results, while the baseline's performance drops very quickly. }

\begin{tabular}{ccc}
\toprule

Data Volume & Baseline  & Ours \\
1           & 48.3 & 52.8 \\
1/2         & 48.0 & 51.5 \\
1/4         & 46.6 & 50.5 \\
1/8         & 41.1 & 49.6 \\
1/16        & 18.2 & 46.9\\

\bottomrule

\end{tabular}

\label{Table:decrease data}
\vspace{+0.2cm}
\end{table}

\begin{table}[t]

\centering
\small

\caption{Comparison different backbone. The results are evaluated on the PASCAL VOC 2012 training set.}

\begin{tabular}{cccccc}
\toprule
Backbone           & Enhanced &Base  & Parameter & Top 1 error  \\ \hline
resnet50           & 49.8 &48.3 & 25.5M     & 23.8              \\
resnet101          & 51.0 &49.7 & 44.5M     & 22.6        
\\

xception          & 50.5 &49.4 & 28.8M     & 21.2        
\\

wide\_resnet50\_2  & 49.7 &48.4 & 68.9M     & 21.4              \\
wide\_resnet101\_2 & 50.8 &49.6 & 126.9M    & 21.1             \\
resnext\_50        & 50.8 &49.1 & 25.0M     & 22.3           \\
resnext\_101       & 52.8 &50.4 & 44.3M     & 20.6       \\

\bottomrule

\end{tabular}

\label{Table:backbone}
\end{table}
\begin{table}[t]

\centering
\small

\caption{Comparison different features combination.}

\begin{tabular}{cccc}
\toprule
block2 & block3 & block4 & mIoU      \\ \hline
\checkmark      &        &        & 50    \\
       & \checkmark      &        & 50.4  \\
       &        & \checkmark      & 50.1  \\
\checkmark      & \checkmark      &        & 48.83 \\
\checkmark      &        & \checkmark      & \textbf{50.8}  \\
       & \checkmark      & \checkmark      & 50.12 \\
\checkmark      & \checkmark      & \checkmark      & 50.3   \\
\bottomrule

\end{tabular}

\label{Table:feature-comparison}
\end{table}
\begin{table}[t]

\centering
\small

\caption{Comparison different probability to discard the object regions. The results are evaluated on the PASCAL VOC 2012 training set.}

\begin{tabular}{cc}
\toprule
Probability & mIoU   \\ \hline
0.1         & 50.1   \\
0.3         & 50.2   \\
0.5         & 50.8   \\
0.7         & 50.5 \\
0.9         & 50.3  \\     
\bottomrule

\end{tabular}
\label{Table:hyper-parameter}
\end{table}

 \begin{figure*}[t]
    \centering
    \includegraphics[width=0.7\linewidth]{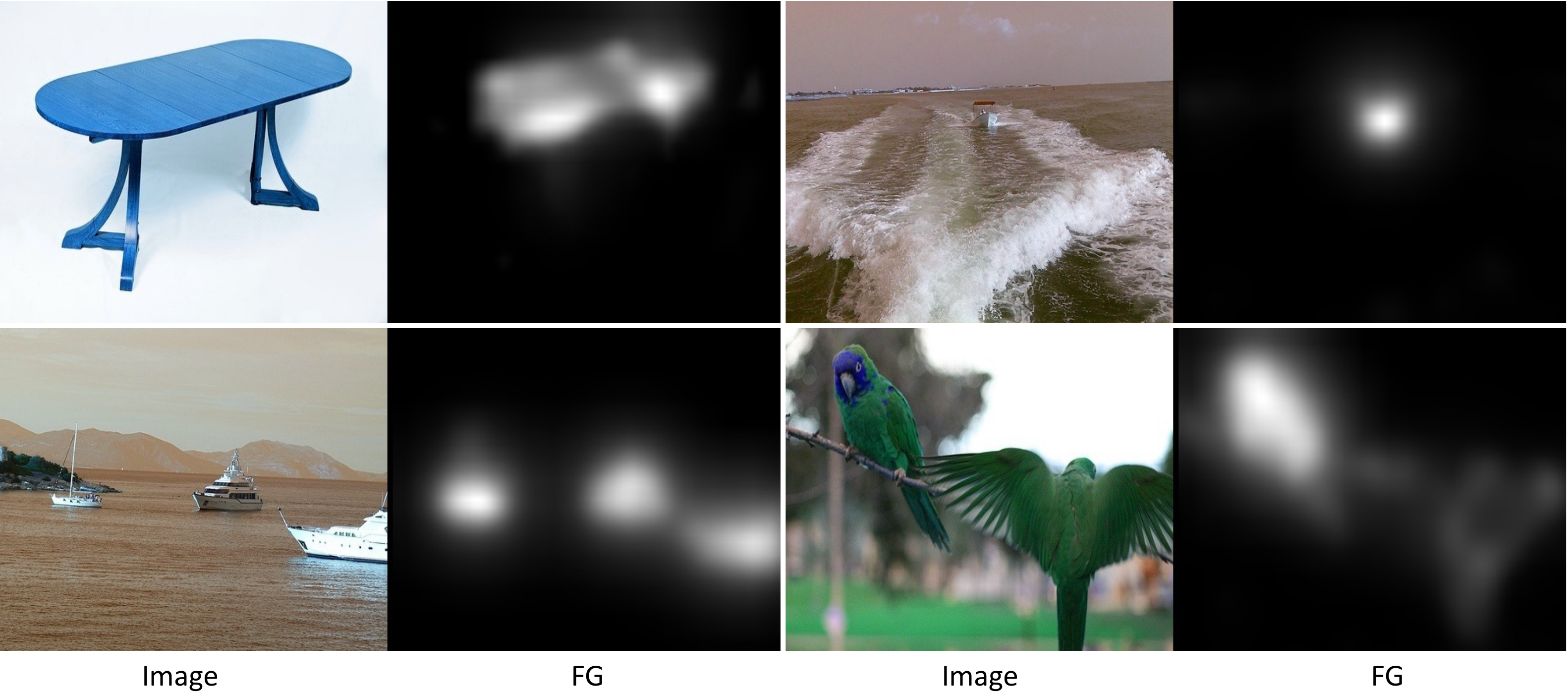}
    \caption{ Visualization examples of the foreground probability maps for PASCAL VOC 2012 dataset. FG denotes the foreground probability maps $\hat{F}$.}
    \label{Figure:foreground}
\end{figure*}

\begin{table}[t]

\setlength{\tabcolsep}{7mm}{
\centering
\small

\caption{The performance of our RPNet with and without Gaussian Filter. GF denotes Gaussian Filter. The results are evaluated on the PASCAL VOC 2012 training set.}

\begin{tabular}{c|c}
\toprule
 Ours w GF  & Ours w/o GF  \\ \hline
50.8    & 50.3  \\
\bottomrule
\end{tabular}

\label{Table:Gauss_filter}
}
\end{table}
\begin{table}[t]
\centering
\small

\caption{The effectiveness of the relations between image pairs.}

\begin{tabular}{cc}
\toprule

method              & mIoU \\ \hline
baseline           & 48.3 \\
prototype with one image & 49.8 \\
prototype with two image & 50.8 \\  
prototype with three image & 51.0 \\  
prototype with four image & 51.1 \\  
\bottomrule

\end{tabular}

\label{Table:two-way}
\end{table}

\begin{table*}[]
\parbox{.45\linewidth}{
\centering
\caption{Comparison of the weakly supervised semantic segmentation methods. With the setting of without any additional supervision, our method outperforms all the previous methods on both validation set and test set. }
  \begin{tabular}{l|c|c|ccc}
  \toprule
    Method & AS & Val & Test\\
          \hline
       FCN-MIL~\cite{Pathak215}  &-  &  25.7 & 24.9 \\
       CCNN~\cite{Pathak15} &-   &  35.3 &35.6 \\
       DCSM~\cite{dcsm}  &-   &  44.1 &45.1 \\
       BFBP~\cite{bfbp} &-  &  46.6   &  48.0 \\
       SEC~\cite{sec}  &- &  50.7 &51.7 \\
       CBTS~\cite{cbts} &-   &  52.8 &53.7 \\
       TPL~\cite{tphase} &- &  53.1 &53.8 \\
       MEFF~\cite{meff} &- &  - &55.6 \\
       PSA~\cite{psa}  &-   &  61.7 &63.7 \\
       IRN~\cite{irn} &- & 63.5 & 64.8 \\ 
       SSDD.~\cite{ssdd} &- & 64.9 &  65.5 \\
    CIAN~\cite{cian} &- & 62.4 &  65.3 \\
      MBMNet~\cite{lwd2020weakly}&- & 66.2 &  67.1 \\
       MCI~\cite{mining-cross} &- & 66.2 &  66.9 \\
       SEAM~\cite{seam} &- & 64.5 &  65.7 \\
       BENet~\cite{benet} &- & 65.7 &  66.6 \\
        CONTA~\cite{conta}  &- & 65.3 &  66.1 \\
        \hline
       \hline
       \textbf{RPNet} \tiny{(Resnet-50 w/o CRF)}      & -         & \textbf{65.1} & \textbf{66.0} \\ 

\textbf{RPNet} \tiny{(Resnet-50)}        & -         & \textbf{66.4} & \textbf{67.2} \\ 

       \textbf{RPNet} \tiny{(Resnet-101 w/o CRF)}      & -         & \textbf{65.4} & \textbf{66.4} \\ 

\textbf{RPNet} \tiny{(Resnet-101)}        & -         & \textbf{66.9} & \textbf{68.0} \\ 

       \textbf{RPNet} \tiny{(Resnet-38 w/o CRF)}      & -         & \textbf{65.3} & \textbf{66.4} \\ 

\textbf{RPNet} \tiny{(Resnet-38)}        & -         & \textbf{66.7} & \textbf{67.8} \\

    \textbf{RPNet} \tiny{(Resnext-50 w/o CRF)}       & -         & \textbf{65.7} & \textbf{66.7} \\ 

    \textbf{RPNet} \tiny{(Resnext-50) }       & -         & \textbf{67.0} & \textbf{68.1} \\ 
    
         \textbf{RPNet} \tiny{(Resnext-101 w/o CRF)}        & -  & \textbf{66.3} & \textbf{66.0} \\
         
     \textbf{RPNet}\tiny{ (Resnext-101) }      & -  & \textbf{68.0} & \textbf{68.2} \\ 
              \bottomrule
  \end{tabular}

\label{Table:comapre-soa-no-ad}
}
\hfill
\parbox{.5\linewidth}{
\centering
\caption{Comparison of the weakly supervised semantic segmentation methods with more additional supervision, our method outperforms all the previous methods on both validation set and test set even we do not use any additional supervision. WV denote Web Video, S denote Saliency Mask, WI denote Web Image, IS denote Instance Image. }

\begin{tabular}{l|c|c|ccc}
  \toprule
    Method & AS & Val & Test\\
          \hline
        MIL-seg~\cite{ped15}  & S+Img & 42.0  & 40.6\\
       MCNN~\cite{mcue}     & WV  &38.1   &  39.8 \\
       AFF~\cite{afss}   & S & 54.3   &  55.5 \\
       STC~\cite{stc}    & S + WI & 49.8   &  51.2 \\
      Oh et al.~\cite{joon17cvpr}   & S &  55.7   &  56.7 \\
       AE-PSL~\cite{erasing}   & S & 55.0   &  55.7 \\
      Hong et al.~\cite{webvideo-seg}  & WV  &  58.1   &   58.7\\
      WebS-i2~\cite{cvpr17web}    & WI & 53.4 &55.3 \\
      DCSP~\cite{dcsp}    & S & 60.8 &61.9 \\
      GAIN~\cite{gain}     & S & 55.3 &56.8 \\
       MDC~\cite{mdc}      & S & 60.4 & 60.8 \\
      MCOF~\cite{mcof}   & S & 60.3 &61.2 \\
      DSRG~\cite{dsrg}  & S & 61.4 &63.2 \\
      Shen et al.~\cite{cvpr18web}   & WI  &  63.0 & 63.9 \\
      SeeNet~\cite{seenet}   & S & 63.1 &62.8 \\
       AISI~\cite{salins}      & IS & 63.6 &64.5 \\
       FickleNet~\cite{ficklenet} & S & 64.9 & 65.3 \\ 
       DSRG+EP.~\cite{DSRGEP} & S & 61.5 & 62.7 \\ 
      Zeng et al.~\cite{zeng_iccv2019} & S & 63.3 & 64.3 \\ 
      OAA+.~\cite{oaa} & S & 65.2 & 66.4 \\ \hline \hline

\textbf{RPNet} \tiny{(Resnet-50 w/o CRF)}      & -         & \textbf{65.1} & \textbf{66.0} \\ 

\textbf{RPNet} \tiny{(Resnet-50)}        & -         & \textbf{66.4} & \textbf{67.2} \\ 

       \textbf{RPNet} \tiny{(Resnet-101 w/o CRF)}      & -         & \textbf{65.4} & \textbf{66.4} \\ 

\textbf{RPNet} \tiny{(Resnet-101)}        & -         & \textbf{66.9} & \textbf{68.0} \\ 

       \textbf{RPNet} \tiny{(Resnet-38 w/o CRF)}      & -         & \textbf{65.3} & \textbf{66.4} \\ 

\textbf{RPNet} \tiny{(Resnet-38)}        & -         & \textbf{66.7} & \textbf{67.8} \\

    \textbf{RPNet} \tiny{(Resnext-50 w/o CRF)}       & -         & \textbf{65.7} & \textbf{66.7} \\ 

    \textbf{RPNet} \tiny{(Resnext-50) }       & -         & \textbf{67.0} & \textbf{68.1} \\ 
    
         \textbf{RPNet} \tiny{(Resnext-101 w/o CRF)}        & -  & \textbf{66.3} & \textbf{66.0} \\
         
     \textbf{RPNet}\tiny{ (Resnext-101) }      & -  & \textbf{68.0} & \textbf{68.2} \\ 
              \bottomrule
  \end{tabular}

\label{table:compare-with-soa-with-ad}
}
\end{table*}
\begin{table}[t]

\setlength{\tabcolsep}{7mm}{
\centering
\small
\caption{Comparison of the weakly supervised semantic segmentation methods with other methods on MS COCO dataset, our RPNet can outperforms all the previous methods.}

\begin{tabular}{l|c}
\toprule
Method & val  \\ \hline
BFBP~\cite{bfbp}   & 20.4 \\
SEC~\cite{sec}   & 22.4 \\
SEAM~\cite{seam}   & 31.9 \\
IRNet~\cite{irn}  & 32.6 \\
CONTA~\cite{conta}  & 33.4 \\ \hline \hline

\textbf{RPNet}  & \textbf{38.6} \\ 
\bottomrule
\end{tabular}

\label{Table:coco-soa}
\vspace{+0.2cm}
}
\end{table}

\begin{equation}
p_n = \frac{\sum_{x,y}^{}f_n(x,y)M_n{(x,y)} }{ \sum_{x,y}^{} M_n{(x,y)}}.
\label{equation:prototype}
\end{equation}
where
\begin{equation}
M_n(x,y) = \begin{cases}
    $Z$, & \text{if $O_n^{'}(x,y) > \beta$},\\
    0, & \text{otherwise}.
  \end{cases} 
\label{equation:masks}
\end{equation} 
and $Z$ follows Bernoulli distribution $X\sim B(x,\beta)$ and is equal to $1$ with a probability $\beta$. For each $f_n$, we produce a binary confidence mask $M_n{} \in \mathbb{R}^{X \times Y \times 1}$ by thresholding its corresponding confident object activation map $O^{'}$ with $\beta$, as in equation \ref{equation:masks}. The feature $f_n$ is then masked by $M_n$ to produce a prototype vector $p_n$ by average pooling at the spatial dimension with normalization, see equation \ref{equation:prototype}. The parameter $\beta$ helps randomly discard some of the activated regions from $M_n$ during training, and the location of the corresponding $f_n$ will be masked. This is inspired by \cite{seenet} to introduce robustness and help find more object regions. The effectiveness of discarding the activated regions is validated with experiments in Sec.~\ref{section:ablative-study}. The complete flow of prototype generation is illustrated in Figure~\ref{Figure:protype-extractor}.

\textbf{Region comparison.}
\label{sec:generate map}
With the region prototype vectors, we try to filter the unconfident object-related regions in a non-parametric metric learning manner. At first, we calculate the similarity maps $s_n  \in \mathbb{R}^{X \times Y \times 1}$ between the feature maps of the last deep CNN block $f_N$ and the region prototype vectors $p_n$, where $n={1...2N}$ at each spatial position; then, we find the maximum similarity and assign every spatial position with a probability to object foreground depending on it. Collecting all the max similarity values as a set, we form the foreground probability maps $\hat{F}$. We follow Oreshkin~\textit{et al.}~\cite{prototipical} to use cosine as our similarity function, but we do not utilize the scale factor. The foreground probability maps $\hat{F} \in \mathbb{R}^{X \times Y \times 1}$ are generated with the following steps:

\begin{equation}
    s_n(x,y) = \textit{cos}(p_n,f_N(x,y)), n=1...2N,
\end{equation}

\begin{equation}
    \hat{F}(x,y) = max(s_n(x,y)),
\end{equation}
where $N$ is the number of feature maps extracted from different blocks of the CNN network. We also show the visualization examples of the foreground probability maps for PASCAL VOC 2012 dataset in Figure~\ref{Figure:foreground}.

We discuss how we address the two flaws mentioned in Sec. \ref{section:motivation}. For flaw 1), object regions that are not activated in one sample by the classification objective can be re-activated by other samples, through searching prototypes with high confidence value $O'$ and high similarity value $s$. Similar prototypes can be found with sufficient diversity of the training set and can re-activate via feature enhancement discussed in the following section. For flaw 2), if a prototype $p_i$ belonging to the background is wrongly extracted, although its confidence scores $O'_i$ is high, it is difficult to find regions that are highly similar to it on other training images due to the diversity assumption. Thus, its information propagation is limited.

\textbf{Feature enhancement by re-weighting the feature maps.}
After we obtain the foreground probability maps $\hat{F}$, we can locate the object more accurately and comprehensively. We extract the feature maps of the last block $f_N(x,y)$ by multiplying with its confidence map $\hat{F}$ (note that there are two $\hat{F}$ and each corresponds to one of the image pairs). As the naive $\hat{F}$ is sparse, we utilize a Gaussian Filter~\cite{gauss} to smooth $\hat{F}$. In particular, we re-weight the original feature representations in the following way:

\begin{equation}
    G(x,y) = \frac{1}{2 \pi \sigma ^{2}} e^{- \frac{x^{2} + y^{2}}{2 \sigma ^{2}}},
\label{2dgaussian}
\end{equation}

\begin{equation}
    \mathbf{f_N(x,y)^{'}} = G(\hat{F}_{(x,y)})\cdot \mathbf{f_N(x,y)},
\end{equation}
where $f_N(x,y)^{'} \in \mathbb{R}^{X \times Y \times D}$ denote the enhanced object activation maps.
With the enhanced object activation maps, we employ an average pooling followed by a $1 \times 1$ convolution layer as the final classifier to predict the classification.

 \begin{figure*}[t]
  \centering
    \includegraphics[width=1\linewidth]{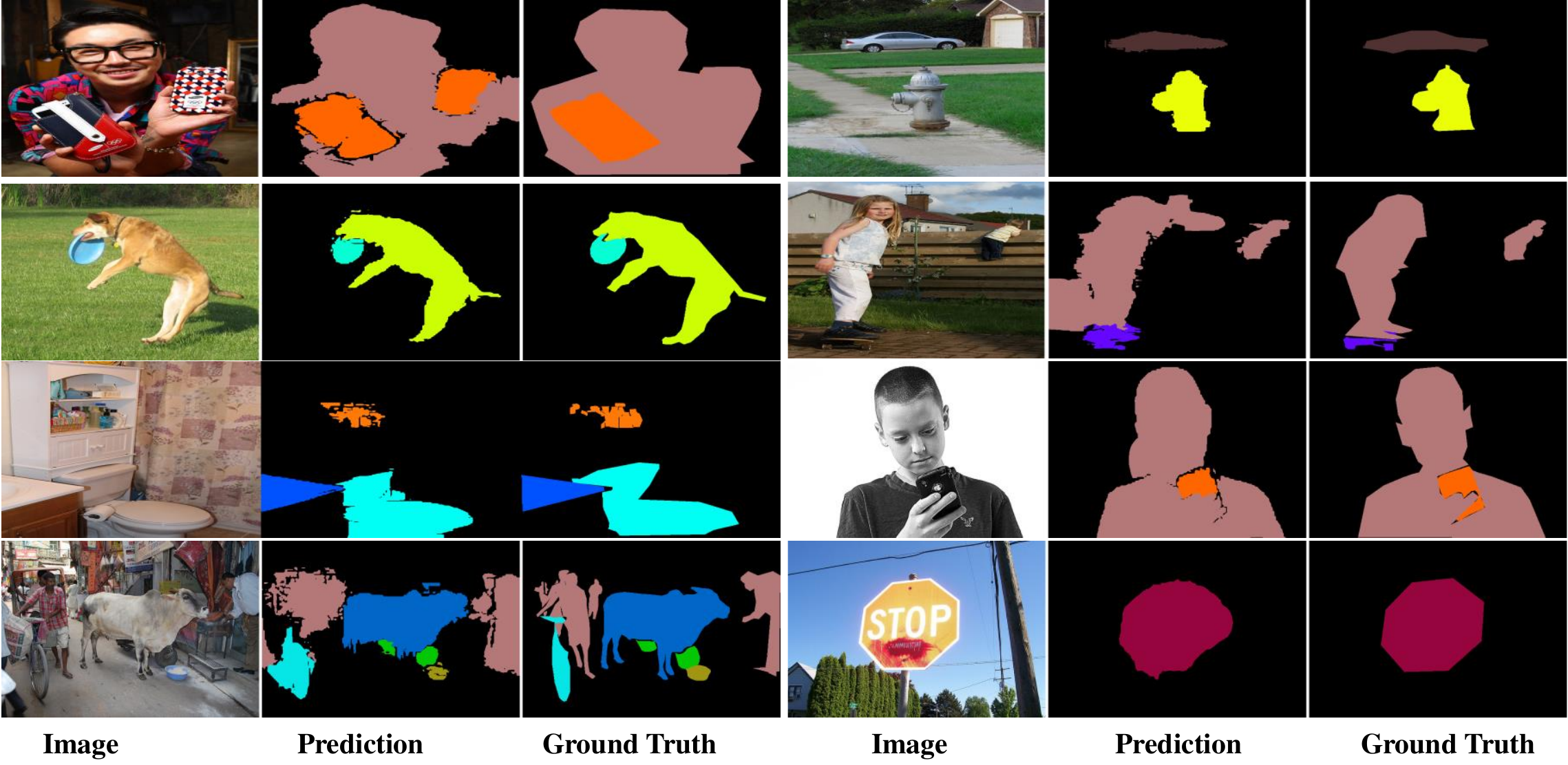}
    
    \caption{Our qualitative examples on the MS COCO.}
    \label{Fig:quality-results-coco}
\vspace{+0.2cm}
\end{figure*}
 \begin{figure*}[t]
  \centering
    \includegraphics[width=1.05\linewidth]{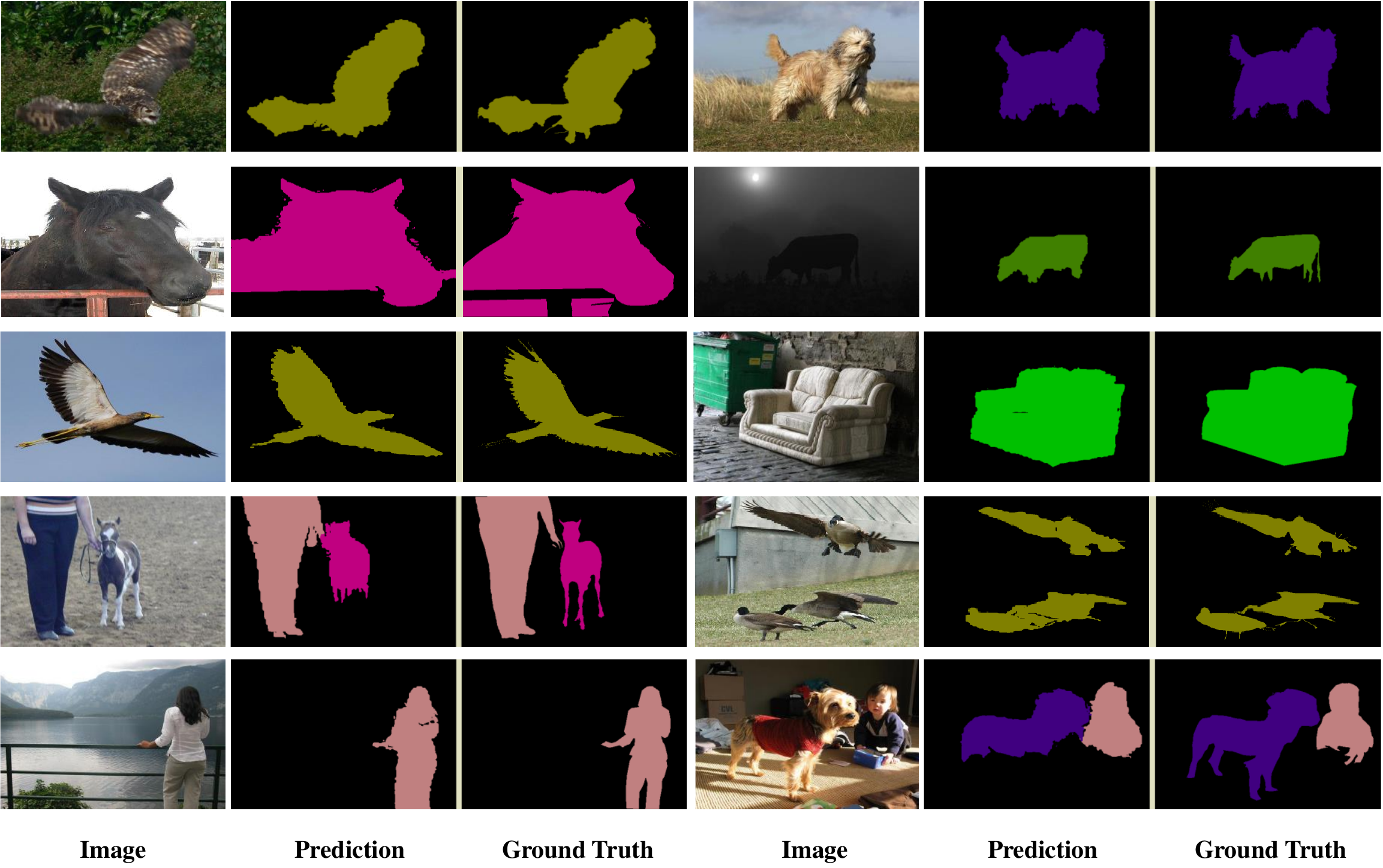}
    
    \caption{Our qualitative examples on the PASCAL VOC 2012.}
    \label{Fig:quality-results}
\vspace{+0.2cm}
\end{figure*}

\begin{table*}[]

\small
\centering
\caption{Detail results on the PASCAL VOC 2012 dataset. Our proposed method outperforms all previous methods on both val set and test set.   }

\begin{subtable}{1\textwidth}
\small
\centering

\resizebox{\columnwidth}{!}
{
\scriptsize
\begin{tabular}[c]{c|*{21}{@{\hspace{0.07cm}}c@{\hspace{0.07cm}}}|cc}
\toprule
methods &\rotatebox{90}{bg}& \rotatebox{90}{aero}&
 \rotatebox{90}{bike}& \rotatebox{90}{bird}& \rotatebox{90}{boat}&
 \rotatebox{90}{bottle}& \rotatebox{90}{bus}& \rotatebox{90}{car}&
 \rotatebox{90}{cat}& \rotatebox{90}{chair}& \rotatebox{90}{cow}&
 \rotatebox{90}{table}& \rotatebox{90}{dog}& \rotatebox{90}{horse}&
 \rotatebox{90}{motor}& \rotatebox{90}{person}& \rotatebox{90}{plant}&
 \rotatebox{90}{sheep}& \rotatebox{90}{sofa}& \rotatebox{90}{train}&
 \rotatebox{90}{tv}& \rotatebox{90}{\textbf{mIoU}} \\ \hline
 
\scriptsize{ RPNet (Resnet-50)} &89.8	& 79.0	& 33.8	& 82.1	& 62.1	& 68.4	& 86.6	& 79.2	&79.1	&30.7	&77.2	&26.4	&73.9	& 78.0	&75.1	&73.8	& 51.1	&80.5	&36.6	&68.1	& 63.7 & 66.4\\

\scriptsize{ RPNet (Resnext-50)} &89.7	& 69.4	& 32.8	& 82.2	& 67.2	& 71.1	& 87.7	& 78.7	&80.3	&32.8	&79.1	&34.7	&76.5	& 80.9	&74.2	&74.8	& 53.2 &80.0	&42.5	&65.2	& 55.0 & 67.0\\

\scriptsize{ RPNet (Resnext-101)} &89.7	& 61.0 & 31.8	& 86.6	& 60.9	& 68.1	& 87.3	& 80.4	&88.4	&32.7	&80.0	&42.7	&83.3	& 81.9	&76.6	&73.7	& 54.1	&84.9	&40.9	&63.7	& 60.1 & 68.0\\

\bottomrule
\end{tabular}
}

\caption{The detail results on PASCAL VOC 2012 {\it val set}. \label{table:val_a}} 
\end{subtable}

\medskip
\begin{subtable}{1\textwidth}
\small
\centering
\resizebox{\columnwidth}{!}{
\scriptsize
\begin{tabular}[c]{c|*{21}{@{\hspace{0.07cm}}c@{\hspace{0.07cm}}}|cc}
\toprule
methods &\rotatebox{90}{bg}& \rotatebox{90}{aero}&
 \rotatebox{90}{bike}& \rotatebox{90}{bird}& \rotatebox{90}{boat}&
 \rotatebox{90}{bottle}& \rotatebox{90}{bus}& \rotatebox{90}{car}&
 \rotatebox{90}{cat}& \rotatebox{90}{chair}& \rotatebox{90}{cow}&
 \rotatebox{90}{table}& \rotatebox{90}{dog}& \rotatebox{90}{horse}&
 \rotatebox{90}{motor}& \rotatebox{90}{person}& \rotatebox{90}{plant}&
 \rotatebox{90}{sheep}& \rotatebox{90}{sofa}& \rotatebox{90}{train}&
 \rotatebox{90}{tv}& \rotatebox{90}{\textbf{mIoU}} \\ \hline

\scriptsize{ RPNet (Resnet-50)} & 90.4 & 79.9
& 35.1
& 83.5
& 58.2
&65.5
& 84.7
& 80.6
&76.9
&30.5
& 78.9
&33.4
&75.6
& 78.3
& 80.2
&74.3
&46.0
& 83.7
&50.3
&64.0
& 60.6
& 67.2\\ 

\scriptsize{ RPNet (Resnext-50)} & 90.3 & 77.5
& 33.5
& 83.1
& 59.7
&68.6
& 86.8
& 81.1
&79.0
&31.5
& 81.1
&40.5
&79.9
& 81.9
& 81.9
&74.9
&49.3
& 85.7
&53.3
&57.8
& 52.9
& 68.1\\ 

\scriptsize{RPNet (Resnext-101)} &90.2 &65.7
&34.5
&88.0
&53.3
&69.5
&87.3
&81.6
&88.3
&32.6
&77.9
&47.0
&85.1
&80.5
&81.4
&74.5
&44.5
&85.6
&52.7
&53.5
&58.7
&68.2\\

\bottomrule

\end{tabular}
}
\caption{The details results on PASCAL VOC 2012 {\it test set}. \label{table:test}}
\end{subtable}
\medskip

\label{Table:details of the results}
\vspace{+0.2cm}
\end{table*}

\subsection{Training Process}
We choose a multi-label soft cross-entropy classification loss $\mathcal{L}_{c}$ to make sure that the activated regions on $f_N^{'}$ still possess discrimination power to the object categories indicated by the image-level labels. 

To solve the problem of incomplete object regions motioned in section~\ref{sec: introduction} and constraint the $f_N^{'}$ from deviating too much from the original feature maps $f_N$, which may cause divergence, we add a self-supervised loss $\mathcal{L}_{s}$ to compare both maps. 

\begin{equation}
 \mathcal{L}=\mathcal{L}_{c}+\lambda\mathcal{L}_{s},
 \label{equation:loss_sum}
\end{equation}

\vspace{-1em}
\begin{multline}
\mathcal{L}_{c}=- \sum_{i=1}^{N}( u[i] * log(\frac{1}{1+exp(-v[i])}) \\ +(1-u[i]) * log(exp(\frac{-v[i]}{1+exp(-v[i])}))),
\end{multline}

 \begin{equation}
 \mathcal{L}_{s}=\sum_{i=1}^{HW}{(\mathbf{f_N'}-\mathbf{f_N})^2}.
  \end{equation}

Here $v$ denotes the predicted probability of class $i$ while $u[i]$ denotes the image level groundtruth of $i_{th}$ class. $f_N$ and $f_N'$ denote the object activation maps and its enhanced ones. We balance the losses with $\lambda$.
\section{Experiment}
\subsection{Dataset and Evaluation Metric}
Following the experiment set-ups in previous works, we train and evaluate our RPNet on the PASCAL VOC 2012~\cite{voc} and MS COCO~\cite{coco14} dataset. The PASCAL VOC2012 training set is extended with images from~\cite{moredata} and finally gets 10582 training images, 1449 validation images, and 1456 testing images. MS-COCO~\cite{coco14} contains 81 classes, 80k training images, and 40k val images. We use image-level class labels only.
We adopt the standard mean Intersection-over-Union(mIoU) as the evaluation metric for all of our experiments.

\subsection{Implementation Details}
All the backbones are pretrained on ImageNet~\cite{imgnet}, and we replace the stride of the last layer from 2 to 1 to maintain the feature map size. The batch size is set to 16.
We apply the random crop, random scale, and random flip to the images for data augmentation during the training process. The learning rate is initially set to be 0.02 with a decrease rate of 0.9 at every iteration with polynomial decay~\cite{parsenet}. We use SGD as the optimizer and train the network for 5 epochs. We set the probability $\beta$ to 0.5 to discard the object activation regions as described in generate region prototypes, Sec. \ref{sec: rpnet} , and set the $\sigma$ to 3 as motioned in feature enhancement, equation~\ref{2dgaussian}, set the $\lambda$ in equation~\ref{equation:loss_sum} to be 10 and set the $\alpha$ to be 0.3 in equation~\ref{equation:masks}. We adopt the boundary refinement method proposed in~\cite{irn} and synthesize the pixel-level pseudo groundtruth to train a DeepLab-LargeFOV~\cite{deeplab} with Resnet-50 backbone as our semantic segmentation model. 
Finally, we follow the previous works~\cite{seenet,sec,ssdd} to utilize denseCRF~\cite{crf} as post-processing to further refine our predictions. 

\subsection{Ablative Analysis} \label{section:ablative-study}
In this section, if not specified, we employ Resnext-50 as the network backbone as default; all experiments are conducted on PASCAL VOC 2012 training set and evaluated with standard mIoU. We choose the ResNext-50 instead of ResNet50 as the backbone because choosing ResNext-50 over Resnet-50 is part of our contribution: each channel group from a grouped convolution layer could learn a unique representation of the data to form the fused result to contain more activation regions, which provides significant improvements.

\textbf{The Suitable Backbone.}
As is shown in Figure~\ref{Figure:different-backbone}, we compare backbones and argue that the grouped convolution achieves a significant positive influence on the weakly supervised segmentation task: each channel group could learn a unique representation of the data \cite{group_convolution} and the fused result is more robust.
We compare different backbone choices in Table~\ref{Table:backbone}. We find that group convolution plays more important roles compared with the number of parameters and network structures.

\textbf{Feature comparison.}
In Table~\ref{Table:feature-comparison}, We compare our model variants that utilize different level features from our backbone encoder to generate the foreground probability maps. 
We experiment with the feature comparison with single block features and combining multiple block features. The best performance is achieved by combining block 2 and block 4. The possible explanation is that block2 corresponds to relatively low-level localization cues but lacks semantic cues while block4 is the opposite. This combination makes up for each other's shortcomings and gives play to their advantages, which leads to the final best performance. The block3 achieves the best performance with a single block is used; the possible explanation is that block3 contains both the low-level cues and the high-level cues consistent with our previous explanation.

\textbf{The hyper-parameter choices.}
We investigate the choices of $\beta$, the possibility to discard a patch of the region during prototype generation. As shown in Table~\ref{Table:hyper-parameter}, the best performance achieves when discarding the patches with the probability of 0.5.

\textbf{The effectiveness of Gaussian Filter.}
We utilize the Gaussian Filter to smooth the naive sparse $\hat{F}$. We validate the effectiveness of the Gaussian Filter in Table~\ref{Table:Gauss_filter}. The Gaussian filter can bring 0.5 mIoU improvement on the VOC 2012 training set. 

\textbf{The effectiveness of cross-image relations.}
To validate the effectiveness of our region prototype and the effectiveness of cross-image relations, we compare our network to a baseline model that does not employ our region prototype mechanism. 
As is shown in Table~\ref{Table:two-way}, when the region prototype vectors only from one image, our methods can bring 0.7 mIoU improvement over the baseline. While utilizing the cross-image relations with two images, our RPNet can bring 1.7 mIoU improvement. The results show that significant improvement when we use one or two images. Due to the GPU memory limitation, we report results of up to 4 images per iteration.

\subsection{Decrease the number of training samples} 
\label{sec:decress_training}
We investigate the robustness of the proposed method on datasets of reduced training samples. We reduce the number of training samples of each object category from PASCAL VOC 2012 and re-train our RPNet, and the baseline \cite{irn}. As shown in Table \ref{Table:decrease data}, the performance baseline method \cite{irn} deteriorates significantly as the number of samples is reduced to $1/16$ to its original number, but our proposed method's quality drops slowly. 
As shown in Figure~\ref{Fig:decrease-data}, decreasing the training image, the baseline method can not identify the object accurately, especially with more background regions that are wrongly activated. This is because the fewer images, the less diversity required from the extracted features for classification. This leads to the activation regions are not precisely the objects yet still discriminative. 
Even with the decrease of the training images, our method still attempts to explore the discriminative regions and the cross-image similarities among these regions. Thus, more falsely activated backgrounds are suppressed. Even if the number of images of each class is decreased to one, the method still performs non-local feature comparison within one image, suppressing as much background as possible.
Since there is more diversity among backgrounds, fewer compared regions during prototype voting resemble the activated background region, and its information propagation is minimized. Our region comparison serves as a voting mechanism to assign them with lower confidence scores, which suppresses the inaccurate background. 

\subsection{Analysis of Pseudo Labels}
The performance of our synthesized segmentation labels is measured with standard mIoU on PASCAL VOC 2012~\cite{voc} and MS COCO~\cite{coco14} training data. As shown in Table~\ref{psedo_mask}, with the same boundary refine method~\cite{irn}, the synthesized segmentation labels with enhanced object activation maps perform better than the base object activation maps and other state-of-the-art methods~\cite{conta} on both datasets. With the same baseline(IRN) and Boundary Refine method, our RPNet performance 9.7 mIoU is higher than the most current state-of-the-art method CONTA~\cite{conta} on the large dataset MS COCO. 

\subsection{Comparison with State-of-the-art Results}
For performance comparison, as shown in Table~\ref{Table:comapre-soa-no-ad} and Table~\ref{Table:coco-soa}, we achieve state-of-the-art performances on both datasets. Our method does not require any additional training information and even outperforms some methods with it, as in \cite{dcsp,gain,dsrg,seenet,salins,cvpr18web}.

Our RPNet outperforms the results of all previous methods with all types of backbone, such as Resnet-50, Resnext-50, and Resnext-101.
We also verify our method with/without the denseCRF post-processing. It is observed that our method achieves the best performance regardless of denseCRF.
 
As shown in Table~\ref{Table:coco-soa}, we evaluate our methods on the MS COCO dataset, which contains more classes and training images; this means the images have more diversity. Our RPNet achieves new state-of-the-art performances; to be noticed, our methods achieve more than 5 mIoU improvement over the other state-of-the-art methods. As shown in Table~\ref{psedo_mask}, our RPNet can generate high-quality pseudos segmentation masks, which achieves 17.5 mIoU improvement over the baseline and 9.7 mIoU higher than the previous state-of-the-art methods. 

We also show the qualitative examples on the PASCAL VOC 2012 in Figure~\ref{Fig:quality-results} and MS COCO in Figure~\ref{Fig:quality-results-coco}. It is particularly interesting to note that in the top left image of Figure~\ref{Fig:quality-results-coco}, the ground truth misses an object(the handphone), but our prediction is able to segment it accurately.
We also provide the detailed results on PASCAL VOC 2012 in Table~\ref{Table:details of the results}.
\section{Conclusion}
We propose a novel framework to generate accurate pixel-level segmentation labels from better object activation maps with image-level labels only. Similar object parts across images are identified via region feature comparison. Object confidence is propagated between regions to discover new object areas while background regions are suppressed. The performances on PASCAL VOC 2012 and MS COCO dataset validate our methods and achieve new state-of-the-art.
\section*{Acknowledgements}
This work is supported by the Delta-NTU Corporate Lab with funding support from Delta Electronics Inc. and the National Research Foundation (NRF) Singapore (SMA-RP10). 
This work is also partly supported by the National Research Foundation Singapore under its AI Singapore Programme (Award Number: AISG-RP-2018-003) and the MOE Tier-1 research grants: RG28/18 (S), RG22/19 (S) and RG95/20, and the National Natural Science Foundation of China (No.61902077).

\ifCLASSOPTIONcaptionsoff
  \newpage
\fi




\bibliographystyle{IEEEtran}
\bibliography{egbib}

\end{document}